\documentclass[nohyperref]{article}

\usepackage{microtype}
\usepackage{graphicx}
\usepackage{subfigure}
\usepackage{booktabs} 

\usepackage{hyperref}

\usepackage[accepted]{icml2022}

\usepackage{amsmath}
\usepackage{amssymb}
\usepackage{mathtools}
\usepackage{amsthm}

\usepackage[capitalize,noabbrev]{cleveref}

\theoremstyle{plain}
\newtheorem{theorem}{Theorem}[section]

\theoremstyle{definition}
\newtheorem{definition}[theorem]{Definition}

\theoremstyle{remark}

\usepackage[textsize=tiny]{todonotes}

\icmltitlerunning{Non-Vacuous Generalisation Bounds for Shallow Neural Networks}

\usepackage{bm}
\newcommand{\E}{\mathbb{E}}

\newcommand{\probmeasure}{\mathcal{M}_1^{+}}

\newcommand{\I}{\bm{1}}
\newcommand{\erf}[1]{\operatorname{erf}\left(#1\right)}
\DeclareMathOperator{\KL}{KL}
\DeclareMathOperator{\smallkl}{kl}
\DeclareMathOperator{\sign}{sign}

\DeclareMathOperator{\argmax}{argmax}

\renewcommand{\Re}{\mathbb{R}}
\renewcommand{\paragraph}[1]{\textbf{#1}}
\renewcommand{\Pr}{\mathbb{P}}
\DeclareMathOperator{\MV}{MV}

\begin{document}

\twocolumn[
\icmltitle{Non-Vacuous Generalisation Bounds for Shallow Neural Networks}



\icmlsetsymbol{equal}{*}

\begin{icmlauthorlist}
\icmlauthor{Felix Biggs}{ucl}
\icmlauthor{Benjamin Guedj}{ucl}
\end{icmlauthorlist}

\icmlaffiliation{ucl}{Centre for Artificial Intelligence and Department of Computer Science, University College London and Inria London, UK}

\icmlcorrespondingauthor{Felix Biggs}{contact@felixbiggs.com}
\icmlcorrespondingauthor{Benjamin Guedj}{b.guedj@ucl.ac.uk}

\icmlkeywords{Machine Learning, ICML}

\vskip 0.3in
]



\printAffiliationsAndNotice{}  

\setlength{\belowcaptionskip}{-10pt}

\begin{abstract}
  We focus on a specific class of shallow neural networks with a single hidden layer, namely those with $L_2$-normalised data and either a sigmoid-shaped Gaussian error function (``erf'') activation or a Gaussian Error Linear Unit (GELU) activation. For these networks, we derive new generalisation bounds through the PAC-Bayesian theory; unlike most existing such bounds they apply to neural networks with \emph{deterministic} rather than randomised parameters.
  Our bounds are empirically non-vacuous when the network is trained with vanilla stochastic gradient descent on MNIST, Fashion-MNIST, and binary classification versions of the above.
\end{abstract}

\section{Introduction}

The study of generalisation properties of deep neural networks is arguably one of the topics gaining most traction in deep learning theory \citep[see, \emph{e.g.}, the recent surveys][]{kawaguchi2020generalization,DBLP:conf/iclr/JiangNMKB20}. In particular, a characterisation of out-of-sample generalisation is essential to understand where trained neural networks are likely to succeed or to fail, as evidenced by the recent NeurIPS 2020 competition "Predicting Generalization in Deep Learning" \citep{DBLP:journals/corr/abs-2012-07976}. One stream of this joint effort, which the present paper contributes to, is dedicated to the study of shallow neural networks, potentially paving the way to insights on deeper architectures.

Despite numerous efforts in the past few years, non-vacuous generalisation bounds for \emph{deterministic} neural networks
with many more parameters than data remain generally elusive. Those few non-vacuous bounds
that exist primarily report bounds for networks with randomised parameters, for
example Gaussian weights, which are re-drawn for every prediction \citep[a
 non-exhaustive list of references would begin
with][]{dziugaite2017computing,DBLP:conf/nips/Dziugaite018,DBLP:conf/nips/NeyshaburBMS17,DBLP:conf/iclr/NeyshaburBS18,DBLP:conf/isit/HellstromD21}, or for compressed versions of the
trained networks \citep{zhou2018nonvacuous}. While these undoubtedly advanced knowledge on generalisation in deep learning theory, this is far from contemporary
practice which generally focuses on deterministic networks obtained directly
through stochastic gradient descent (SGD), as we do.

The PAC-Bayesian theory (we refer to the recent \citealp{guedj2019primer} and \citealp{alquier2021userfriendly} for a gentle introduction) is thus far the only framework within which
non-vacuous bounds have been provided for networks trained on common classification tasks.
Given its focus on randomised or ``Gibbs'' predictors, the aforementioned lack of results for
deterministic networks is unsurprising. However, the framework is not limited to
such results: one area within PAC-Bayes where deterministic predictors are often
considered lies in a range of results for the ``majority vote'', or the expected
overall prediction of randomised predictors, which is itself deterministic.

Computing the average output of deep neural networks with randomised parameters
is generally intractable: therefore most such works have focused on cases where the
average output is simple to compute, as for example when considering linear predictors. Here, building on ideas from \citet{DBLP:journals/corr/abs-2107-03955}, we show
that provided our predictor structure factorises in a particular way, more
complex majority votes can be constructed. In particular, we give formulations for randomised predictors whose majority vote can be expressed as a deterministic single-hidden-layer neural network.
Through this, we obtain classification bounds
for these \emph{deterministic} predictors that are non-vacuous on the celebrated baselines MNIST \citep{726791}, Fashion-MNIST \citep{DBLP:journals/corr/abs-1708-07747}, and binarised versions of the above. We believe these are the first such results.

Our work fundamentally relates to the question: what kind of properties or
structures in a trained network indicate likely generalisation to unseen data?
It has been shown by \citet{DBLP:conf/iclr/ZhangBHRV17} that neural networks
trained by SGD can perfectly overfit large datasets with randomised labels,
which would indicate a lack of capacity control, while simultaneously
generalising well in a variety of scenarios. Thus, clearly any certification of
generalisation must involve extracting additional information other than the
train loss---for example, the specific final network chosen by SGD. How do the
final parameters of a neural network trained on an ``easy'' data distribution as
opposed to a pathological (\emph{e.g.}, randomised label) one differ? A common
answer to this has involved the return of capacity control and the norms of the
weight matrices, often measured as a distance to the initialisation \citep[as
done, \emph{e.g.},
in][]{dziugaite2017computing,DBLP:conf/nips/BartlettFT17,DBLP:conf/iclr/NeyshaburBS18}.

We suggest, following insights from \citet{DBLP:journals/corr/abs-2006-10929},
that a better answer lies in utilising the empirically-observed stability of SGD
on easy datasets. We give bounds that are tightest when a secondary run of SGD
on some subset of the training set gives final weights that are close to the
full-dataset derived weights. This idea combines naturally in the PAC-Bayes
framework with the requirement of perturbation-robustness of the
weights---related to the idea of flat-minima
\citep{DBLP:conf/colt/HintonC93,DBLP:journals/neco/HochreiterS97a}---to
normalise the distances between the two runs. By leveraging this
commonly-observed empirical form of stability we effectively incorporate
information about the inherent easiness of the dataset and how adapted our
neural network architecture is to it. Although it is a deep and interesting
theoretical question as to when and why such stability occurs under SGD, we
believe that by making the link to generalisation explicit we solve some of the
puzzle.

\paragraph{Setting.} We consider \(D\)-class classification on a set
\(\mathcal{X} \subset \Re^{d}\) with ``score-output'' predictors returning values in
\(\mathcal{\hat{Y}} \subset \Re^D\) with multi-class label space \(\mathcal{Y} = [D]\), or in \(\mathcal{\hat{Y}} = \Re\) with binary label space \(\mathcal{Y} = \{+1, -1\}\). The prediction is the argmaximum or sign of the output and the
misclassification loss is defined as
\(\ell(f(x), y) = \I\{\argmax_{k \in [D]} f(x)[k] \ne y\}\) or \(\ell(f(x), y) = \I\{y f(x) \le 0\}\) respectively. It is will prove useful that scaling does not enter into these losses
and thus the outputs of classifiers can be arbitrarily re-scaled by \(c > 0\)
without affecting the predictions. We write
\(L(f) := \E_{(x, y) \sim \mathcal{D}} \ell(f(x), y)\) and
\(\hat{L}(f) := m^{-1} \sum_{(x, y) \in S} \ell(f(x), y)\) for the risk and
empirical risk of the predictors with respect to data distribution
\(\mathcal{D}\) and i.i.d. $m$-sized sample \(S \sim \mathcal{D}^m\).

\paragraph{Overview of our contributions.} We derive generalisation bounds for a single-hidden-layer neural network \(F_{U, V}\) with first
and second layer weights \(U\) and \(V\) respectively taking the form
\[ F_{U, V}(x) = V \, \phi\left( \beta \frac{Ux}{\|x\|_2}\right)\] with \(\phi\)
being an element-wise activation. If the data is normalised to have
\(\|x\|_2 = \beta\) these are simply equivalent to one-hidden-layer neural
networks with activation \(\phi\) and the given data norm.
We provide high-probability bounds on \(L(F_{U, V})\) of the approximate form
\[ 2 \E_{f \sim Q} \hat{L}(f) + \mathcal{O}\left(\frac{\beta \|U - U^n\|_F + \|V - V^n\|_F}{\sqrt{m - n}}\right), \]
where \(Q\) is a distribution over predictors \(f\), which depends on \(U\) and
\(V\) but does not necessarily take the form of a neural network. The
construction of this randomised proxy \(Q\) is central to our PAC-Bayes derived
proof methods.
The bounds hold uniformly over any choice of weight matrices,
but for many choices the bounds obtained will be vacuous; what is interesting is that they are non-vacuous for SGD-derived solutions on some real-world datasets.
\(U^n\) and \(V^n\) are matrices constructed using some subset
\(n < m\) of the data. Since we consider SGD-derived weights, we can leverage
the empirical stability of this training method (through an idea introduced by
\citealp{DBLP:journals/corr/abs-2006-10929}) to construct \(U^n, V^n\) which are
quite close to the final true SGD-derived weights \(U, V\), essentially by
training a prior on the \(n\)-sized subset in the same way.

\paragraph{Outline.} In \Cref{section:background-related} we give an overview of
results from previous works which we use. In \Cref{section:binary} we give
a bound on the generalisation error of binary classification SHEL networks, which are single hidden layer networks with ``erf'' activations.
In \Cref{section:multiclass} we extend to multi-class classification using a simple assumption, giving a general formulation as well as results for ``erf''- and GELU-activated networks.
In
\Cref{section:results} we discuss our experimental setting and give our
numerical results, which we discuss along with future work in
\Cref{section:discussion}.

\section{Background and Related Work}\label{section:background-related}

\paragraph{PAC-Bayesian bounds.} Originated by
\citet{McAllester1998,McAllester1999}, these generally consider the expected
loss or Gibbs risk \(L(Q) := \E_{f \sim Q}L(f)\) and analogously for the
empirical risk, where \(Q \in \probmeasure(\mathcal{F})\) (with \(\probmeasure(\mathcal{A})\) denoting the set of measures on \(\mathcal{A}\)) is a distribution over
randomised predictors \(f \in \mathcal{F}\). The high-probability bounds take
the rough form (although numerous variations using variance terms or attaining
fast rates also exist -- see the aforecited \citealp{guedj2019primer} and \citealp{alquier2021userfriendly} for a survey)
\begin{equation}\label{eq:general-pac-bayes}
L(Q) \le \hat{L}(Q) + \mathcal{O}\left(\sqrt{\frac{\KL(Q, P) + \log(1/\delta)}{m}}\right)
\end{equation}
holding with at least \(1{-}\delta\) probability over the draw of the dataset.
Here \(\KL(Q, P)\) is the Kullback-Leibler divergence and
\(P \in \probmeasure(\mathcal{F})\) is the PAC-Bayesian ``prior'' distribution,
which must be chosen in a data-independent way (but is not subject to the same
requirements as a standard Bayesian prior for the validity of the method). This
bound holds over all ``posterior'' distributions \(Q\), but a poor choice (for
example, one over-concentrated on a single predictor) will lead to a vacuous
bound. We note in particular the following, which we use to prove our main results.

\begin{theorem}\label{theorem:seeger}  \emph{\citet{langford01boundsfor},
    \citet{DBLP:journals/corr/cs-LG-0411099}}. Given data distribution
  \(\mathcal{D}\), \(m \in \mathbb{N}^{+}\), prior
  \(P \in \probmeasure(\mathcal{F})\), and \(\delta \in (0, 1)\), with probability
  \(\ge 1 - \delta\) over \(S \sim D^{m}\), for all \(Q \in \probmeasure(\mathcal{H})\)
  \begin{equation*}\label{eq:seeger-bound}
    L(Q) \le \smallkl^{-1}\left(\hat{L}(Q), \; \frac{1}{m} \left( \KL(Q, P) + \log\frac{2\sqrt{m}}{\delta} \right) \right)
  \end{equation*}
  where \(\smallkl^{-1}(u, c) := \sup \{v \in [0, 1] : \smallkl(u, v) \le c\}\) and
  \(\smallkl(q : p) := q \log(q/p) + (1-q)\log((1-q)/(1-p))\).
\end{theorem}
We note the relaxation \(\smallkl^{-1}(u, c) \le u + \sqrt{c/2}\) which gives an idea of the behaviour of \Cref{theorem:seeger}; however in the case of \(u\) close to \(0\) the original formulation is considerably tighter.

\paragraph{Data-Dependent Priors.} A careful choice of the prior is essential to the
production of sharp PAC-Bayesian results. A variety of works going back to
\citet{DBLP:conf/nips/AmbroladzePS06} and
\citet{DBLP:journals/jmlr/Parrado-HernandezASS12} \citep[and further developed
by][among others]{DBLP:conf/nips/Dziugaite018,DBLP:journals/corr/abs-2006-10929,
  DBLP:conf/nips/RivasplataSSPS18,perezortiz2021learning,JMLR:v22:20-879} have
considered dividing the training sample into two parts, one to learn the prior
and another to evaluate the bound. Formally, we divide
\(S = S^{\operatorname{prior}} \cup S^{\operatorname{bnd}}\) and use
\(S^{\operatorname{prior}}\) to learn a prior \(P^n\) where
\(n = |S^{\operatorname{prior}}|\), then apply the PAC-Bayesian bound using
sample \(S^{\operatorname{bnd}}\) to a posterior \(Q\) learned on the entirety
of \(S\). The resulting bound replaces \(\hat{L}\) by
\(\hat{L}_{\operatorname{bnd}}\), \(P\) by the data-dependent \(P^n\), and
\(m\) by \(m - n = |S^{\operatorname{bnd}}|\); thus the KL complexity term may
be reduced at the cost of a smaller dataset to apply the bound to.

\citet{DBLP:journals/corr/abs-2006-10929} used this when considering training
neural networks by constructing a so-called ``coupled'' prior \(P^n\) which is
trained in the same way from the same initialisation as the posterior \(Q\) by
stochastic gradient descent with the first \(n\) examples from the training set
forming one epoch. Due to the stability of gradient descent, the weights of
\(P^n\) and \(Q\) evolve along similar trajectories; thus stability of the
training algorithm is leveraged to tighten bounds without explicit stability
results being required (and we do not study the conditions under which SGD
provides such solutions). In many ways this can be seen as an extension of
previous work such as \citet{dziugaite2017computing} relating generalisation to
the distance from initialisation rather than total weight norms.

\paragraph{Majority Votes.} Since PAC-Bayesian bounds of the form in \eqref{eq:general-pac-bayes} and \Cref{theorem:seeger} generally
consider the risk of randomised predictors, a natural question is whether
prediction accuracy can be improved by ``voting'' many independently drawn
predictions; such a majority vote predictor takes the deterministic form
\(\MV_Q(x) := \argmax_k \Pr_{f \sim Q}(\argmax f(x) = k)\).
Several strategies have been devised to obtain bounds for these predictors via
PAC-Bayesian theorems, with the simplest (and often most successful) being the
unattributed first-order bound
\(\ell(\MV_Q(x), y) \le 2 \E_{f \sim Q} \ell(f(x), y)\) valid for all \((x, y)\),
called the ``folk theorem'' by \citet{langford2003pac} and the \emph{first-order} bound elsewhere. This can be substituted
directly into PAC-Bayesian theorems such as \Cref{theorem:seeger} above to
obtain bounds for the majority vote at a de-randomisation cost of a factor of two. This is
the result we use, since across a variety of preliminary experiments we found
other strategies including the tandem bound of
\citet{DBLP:conf/nips/MasegosaLIS20} and the C-bound of
\citet{DBLP:conf/nips/LacasseLMGU06} were uniformly worse, as also discussed by \citet{zantedeschi2021learning}.

\paragraph{Gaussian Sign Aggregation.} To exploit the useful relationship above,
\citet{germainPACBayesianLearningLinear2009} considered aggregating a kind of
linear prediction function of the form \(f(x) = \sign(w \cdot x)\) with
\(w \sim Q = N(u, I)\). In this case the aggregation can be stated in closed
form using the Gaussian error function ``erf'' as
\begin{equation}\label{eq:erfaggregation}
\E_{w \sim N(u, I)} \sign(w \cdot x) = \erf{\frac{u \cdot x}{\sqrt{2} \|x\|_2}}.
\end{equation}
This closed-form relationship has been used since by
\citet{NIPS2019_8911} and \citet{DBLP:journals/entropy/BiggsG21} in a PAC-Bayesian context
for neural networks with sign activation functions and Gaussian weights;
\citet{DBLP:journals/corr/abs-2107-03955} used it to derive a generalisation
bound for SHEL (single hidden erf layer) networks, which have a single hidden
layer with erf activation function. We will consider deriving a different
PAC-Bayesian bound for this same situation and develop this method further in
this work.

\paragraph{Other Approaches.} A wide variety of other works have derived
generalisation bounds for deterministic neural networks without randomisation.
We note in particular the important works of
\citet{DBLP:conf/nips/BartlettFT17}, \citet{DBLP:conf/nips/NeyshaburBMS17}
(using PAC-Bayesian ideas in their proofs) and
\citet{DBLP:conf/icml/Arora0NZ18}, but contrary to us, they do not provide
empirically non-vacuous bounds. \citet{DBLP:conf/iclr/NagarajanK19} de-randomise
PAC-Bayesian bounds by leveraging the notion of noise-resilience (how much the
training loss of the network changes with noise injected into the parameters),
but they note that in practice their bound would be numerically large. Many of
these approaches utilise uniform convergence, which may lead to shortcomings as
discussed at length by \citet{DBLP:conf/nips/NagarajanK19}; we emphasise that
the bounds we give are non-uniform and avoid these shortcomings. Finally, we
also highlight the works of
\citet{DBLP:conf/colt/NeyshaburTS15,DBLP:conf/iclr/NeyshaburLBLS19} which
specifically consider single-hidden-layer networks as we do -- as in the recent study from \citet{tinsi2021risk}. Overall we
emphasise that, to the best of our knowledge, all existing bounds for deterministic
networks are vacuous when networks are trained on real-world data.

\section{Binary SHEL Network}\label{section:binary}

We begin by giving a bound for binary classification by a single hidden layer
neural network with error function (``erf'') activation. Binary classification
takes \(\mathcal{Y} = \{+1, -1\}\), with prediction the sign of the prediction
function. The specific network takes the following form with output dimension
\(D = 1\). Although the \(\operatorname{erf}\) activation function is not a commonly-used one, it
is very close in value to the more common \(\operatorname{tanh}\) activation. It can also be
rescaled to a Gaussian CDF activation, which is again very close to the
classical sigmoid activation (and is itself the CDF of the probit distribution).

\begin{definition}\emph{SHEL Network.} \citep{DBLP:journals/corr/abs-2107-03955}
  For \(U \in \Re^{K \times d}\), \(V \in \Re^{K \times D}\), and \(\beta > 0\),
  a \(\beta\)-normalised single hidden erf layer (SHEL) network is defined by
  \[ F^{\operatorname{erf}}_{U, V}(x) := V \cdot \erf{\beta \frac{Ux}{\|x\|_2}}.\]
\end{definition}

The above is a single-hidden-layer network with a first normalisation layer, or
if the data is already normalised the overall scaling \(\|x\|_2\) can be
absorbed into the \(\beta\) parameter. This parameter \(\beta\) could easily be
absorbed into the matrix \(U\) and mainly has the effect of scaling the relative
learning rate for \(U\) versus \(V\) when training by gradient descent, as shown
by looking at \(\frac{\partial}{\partial U} F^{\operatorname{erf}}_{U, V}(x)\), something which would normally be affected
by the scaling of data. A higher \(\beta\) means more ``feature learning'' takes
place as \(U\) has a relatively larger learning rate.

For binary classification, the majority vote of distribution \(Q\) is
\(\MV_Q(x) = \sign(\E_{f \sim Q} \sign(f(x)))\). By expressing the
(binary classification) SHEL network directly as the majority vote of a
randomised prediction function, we can prove a PAC-Bayesian generalisation bound
on its error using the first-order bound. The misclassification error of the
randomised function can further be stated in closed form using the Binomial
cumulative distribution function (CDF), giving rise to a bound where the
distribution \(Q\) does not appear directly.

\begin{theorem}\label{theorem:binary}
  In the binary setting, fix prior parameters
  \(u^0_1, \dots, u^0_K \in \Re^{d}, v^0 \in \Re^K\),
  \(T \in \mathbb{N}^+\), \(\beta > 0\), and data distribution \(\mathcal{D}\).
  For \(\delta \in (0, 1)\), with probability at least \(1 - \delta\) under the sample
  \(S \sim \mathcal{D}^m\), simultaneously for any \(U \in \Re^{K \times d}, v \in \Re^K\),
  \begin{equation*}
    L(F^{\operatorname{erf}}_{U, v}) \le 2 \, \smallkl^{-1}\left(\hat{L}(Q^{\otimes T}),\; \frac{T \kappa + \log\frac{2\sqrt{m}}{\delta}}{m} \right).
  \end{equation*}
  Here \(F^{\operatorname{erf}}_{U, v}\) is a SHEL network with \(\beta\)-normalised activation,
  \[ \kappa := \sum_{k=1}^K \frac{|v_k|}{\|v\|_1} \left( \beta^2 \|u_k - u_k^0\|_2^2 + \log \left(2 \frac{|v_k|/\|v\|_1}{|v^0_k|/\|v^0\|_1}\right) \right)\]
  and
  \[ \hat{L}(Q^{\otimes T}) := \frac{1}{m} \sum_{(x, y) \in S} \operatorname{Bin}\left(\frac{T}{2}; \, T, \frac12 \left( 1 + \frac{yF_{U, v}(x)}{\|v\|_1}\right)\right), \]
  for \(\operatorname{Bin}(k; r, p)\) the CDF of a Binomial distribution with parameters \(r, p\).
\end{theorem}

\section{Multi-class Networks}\label{section:multiclass}

We now go further and show that various single-hidden-layer multi-class neural
networks can also be expressed as the expectation of randomised predictors. We
show specific results for multi-class SHEL networks as well as GELU-activation
\citep{DBLP:journals/corr/HendrycksG16} networks as defined below. We also give
a more general form of the result as a aggregation of individual aggregated
predictors which allows these results to be extended further.

We make a simple assumption based on the first-order bound to extend PAC-Bayesian bounds to this case.
This is necessary because under
certain choices of PAC-Bayes posterior \(Q\), the \emph{majority vote} does not give
the same prediction as the \emph{expected vote} as was the case in \Cref{section:binary},
i.e. there exist \(Q\) such that
\(\argmax_k \E_{f \sim Q} f(x)[k] \ne \MV_Q(x)\) at certain adversary-chosen values of \(x\).
Thus we assume that \(L(\E_{f \sim Q} f(x)) \le 2 \E_{f \sim Q}L(f)\), (denoted
\(\star\)), which follows from the first order bound in the case
\(\E_{Q}f(x) \approx \MV_Q(x)\), which we later verify empirically.

\subsection{SHEL Networks}

Here we give a generalisation bound for a multi-class variant of the SHEL network
using the above assumption. The proof is slightly different from the binary case, but still relies on
the useful fact that the SHEL network can be written as the expectation of a
randomised predictor. This predictor however takes a slightly different form to
that in the binary case.

\begin{theorem}\label{theorem:shel-bnd}
  In the multi-class setting, fix prior parameters \(U^n \in \Re^{K \times d}\) and
  \(V^n \in \Re^{D \times K}\), \(\sigma_V > 0\), \(\beta > 0\), and data distribution \(\mathcal{D}\). For
  \(\delta \in (0, 1)\), with probability at least \(1 - \delta\) under the sample
  \(S \sim \mathcal{D}^m\), simultaneously for any
  \(U \in \Re^{K \times d}, V \in \Re^{D \times K}\) such that assumption (\(\star\)) is satisfied,
  \begin{equation*}\label{eq:shel-bound}
    L(F^{\operatorname{erf}}_{U, V}) \le 2 \, \smallkl^{-1}\left(\hat{L}(Q) ,\; \frac{\kappa + \log\frac{2\sqrt{m}}{\delta}}{m} \right).
  \end{equation*}
  Here \(F^{\operatorname{erf}}_{U, V}\) is a SHEL network with \(\beta\)-normalised activation,
  \[ \kappa := \beta^2 \|U-U^0\|_F^2 + \frac{\|V - V^0\|_F^2}{2\sigma_V^2}, \]
  and
  \[ \hat{L}(Q) := \frac{1}{m} \sum_{(x, y) \in S} \Pr\left\{ \argmax \left[W_2 \sign(W_1 x)\right] \ne y \right\}, \]
  with the probability over draws of
  \(\operatorname{vec}(W_2) \sim N(\operatorname{vec}(V), \sigma_V^2 I), \operatorname{vec}(W_1) \sim N(\operatorname{vec}(U), \frac12\beta^{-2} I) \).
  Note that \(\operatorname{vec}\) is the vectorisation operator and \(\sign\) is
  applied element-wise.
\end{theorem}

\paragraph{Differences to \citet{DBLP:journals/corr/abs-2107-03955}.} In their Theorem
5, \citet{DBLP:journals/corr/abs-2107-03955} give a bound for generalisation in
SHEL networks, with \(L(F^{\operatorname{erf}}_{U, V})\) upper bounded under similar conditions to
\Cref{theorem:shel-bnd} by
\begin{equation*}
\hat{L}^{\gamma}(F^{\operatorname{erf}}_{U, V}) + \tilde{O}\left( \frac{\sqrt{K}}{\gamma \sqrt{m}} \left( V_{\infty}\|U - U^0\|_F + \|V\|_F \right)  \right),
\end{equation*}
where
\(\hat{L}^{\gamma}(g) = m^{-1} |\{ (x, y) \in S : g(x)[y] - \max_{k \ne y}g(x)[k] \le \gamma \}|\),
the proportion of \(\gamma\)-margin errors in the training set, and
\(V_{\infty} := \max_{ij}|V_{ij}|\). Thus a margin loss of the actual predictor
used rather than a stochastic one appears. A tighter formulation more similar to
\Cref{eq:shel-bound} is also given in an appendix and the bound could be
similarly adapted to a data-dependent prior.

The derivation of the bound is quite different from ours, relying on a quite
differently-constructed randomised version of \(Q\) (which is however
constructed to have mean \(F^{\operatorname{erf}}_{U, V}\)), and a de-randomisation procedure relying
on margins and concentration rather than a majority vote bound. Both the form of
\(Q\) used and the de-randomisation step lead to issues which we have addressed
through our alternative formulation of \(Q\) and a majority vote bound:
de-randomisation requires a very low variance \(Q\), leading to the
\(\sqrt{K}/\gamma\) term in the bound, which is empirically very large for low
margin losses. Thus as demonstrated in their experiments, the big-O term
increases with widening networks. Finally we note the most important distinction
to our work: contrary to the present work, \citet{DBLP:journals/corr/abs-2107-03955} do not obtain non-vacuous bounds in practice.

\subsection{GELU Networks}\label{section:gelunet}

The Gaussian Error Linear Unit is a
commonly-used alternative to the ReLU activation defined by
\(\operatorname{GELU}(t) := \Phi(t) \, t\) where \(\Phi(t)\) is the standard
normal CDF. Far from the origin, the \(\Phi(t)\) is saturated at zero or one so
it looks much like a smoothed ReLU or SWISH activation (defined by
\citealp{DBLP:conf/iclr/RamachandranZL18} as \(x / (1 + e^{-cx})\) for some
\(c > 0\)). It was introduced to lend a more probabilistic interpretation to
activation functions, and fold in ideas of regularisation by effectively
averaging the output of adaptive dropout \citep{DBLP:conf/nips/BaF13}; its wide
use reflects excellent empirical results in a wide variety of settings.

\begin{definition}\emph{GELU Network.}
  For \(U \in \Re^{K \times d}\), \(V \in \Re^{K \times D}\), and \(\beta > 0\),
  a \(\beta\)-normalised single hidden layer GELU network is defined by
  \[ F^{\operatorname{GELU}}_{U, V}(x) := V \cdot \operatorname{GELU}\left(\beta \frac{Ux}{\|x\|_2}\right) \]
    where \(\operatorname{GELU}(t) := \Phi(t) \, t\).
\end{definition}

\begin{theorem}\label{theorem:gelu-bnd}

  In the multi-class setting, fix prior parameters \(U^n \in \Re^{K \times d}\) and
  \(V^n \in \Re^{D \times K}\), \(\sigma_V > 0\), \(\sigma_U > 0\) \(\beta > 0\), and data distribution \(\mathcal{D}\). For
  \(\delta \in (0, 1)\), with probability at least \(1 - \delta\) under the sample
  \(S \sim \mathcal{D}^m\), simultaneously for any
  \(U \in \Re^{K \times d}, V \in \Re^{D \times K}\) such that assumption (\(\star\)) is satisfied,
  \begin{equation}\label{eq:linear-binary-bound}
    L(F^{\operatorname{GELU}}_{U, V}) \le 2 \, \smallkl^{-1}\left(\hat{L}(Q) ,\; \frac{\kappa + \log\frac{2\sqrt{m}}{\delta}}{m} \right).
  \end{equation}
  Here \(F^{\operatorname{GELU}}_{U, V}\) is a single-hidden-layer GELU network with
  \(\beta\)-normalised activation,
  \[ \kappa := \left(\beta^2 + \frac{1}{\sigma_U^2}\right) \frac{\|U-U^0\|_F^2}{2} + \frac{\|V - V^0\|_F^2}{2\sigma_V^2}, \]
  and \(\hat{L}(Q)\) is
  \[ \frac{1}{m} \sum_{(x, y) \in S} \Pr\left\{ \argmax \left[W_2 (\I_{W_1 x} \otimes (W_1' x))\right] \ne y \right\}, \]
  with the probability is over draws of
  \(\operatorname{vec}(W_2) \sim N(\operatorname{vec}(V), \sigma_V^2 I), \operatorname{vec}(W_1), \sim N(\operatorname{vec}(U), \beta^{-2} I) \) and \(\operatorname{vec}(W_1') \sim N(\operatorname{vec}(V), \sigma_U^2 I)\).
  Here \(\operatorname{vec}\) is the vectorisation operator and the indicator
  function \(\I_y\) is applied element-wise.
\end{theorem}

Although the proof method for \Cref{theorem:gelu-bnd} and the considerations
around the hyper-parameter \(\beta\) are the same as for \Cref{theorem:shel-bnd}
and SHEL networks, one notable difference is the inclusion of the \(\sigma_U\)
parameter. When this is very small, the stochastic predictions are effectively
just a linear two-layer network with adaptive dropout providing the
non-linearity. The ability to adjust the variability of the stochastic network
hidden layer and thus \(\hat{L}(Q)\) is a major advantage over the SHEL
network; in SHEL networks this variability can only be changed through
\(\beta\), which is a fixed parameter related to the deterministic network, not
just a quantity appearing only in the bound.

\subsection{General Form}

Both of the above bounds can effectively be derived from the same formulation,
as both take the form
\begin{equation}\label{eq:general_form_estimator}
F(x) := \E_{f \sim Q} f(x) = \sum_{k=1}^K v_k H_k(x),
\end{equation}
where \(v_k \in \Re^D\) are the column vectors of a matrix
\(V \in \Re^{D \times K}\) and \(H_k: \mathcal{X} \to \Re\) is itself a
predictor of a form expressible as the expectation of another predictor. This means that there exists
a distribution on functions \(Q^k \in \probmeasure(\mathcal{F}^k)\) such that for each
\(x \in \mathcal{X}\), \(H_k(x) = \E_{h \sim Q^k}[h(x)]\). The bound on the generalisation of such predictors takes essentially the same form those given in the rest of this section.

\begin{theorem}\label{theorem:main-general}
  Fix a set of priors \(P^k \in \probmeasure(\mathcal{F}^k)\) for \(k \in [K]\), a prior weight matrix \(V^0 \in \Re^{D \times K}\), \(\sigma_V > 0\),
  \(\delta \in (0, 1)\). With
  probability at least \(1 - \delta\) under the sample \(S \sim \mathcal{D}^m\) simultaneously for any
  \(V \in \Re^{D \times K}\) and set of \(Q^k \in \probmeasure(\mathcal{F}^k)\)
  such that assumption (\(\star\)) holds,
  \begin{equation}
    L(F) \le 2 \, \smallkl^{-1}\left(\hat{L}(Q),\; \frac{\kappa + \log\frac{2\sqrt{m}}{\delta}}{m} \right)
  \end{equation}
  where \(F\) is the deterministic predictor given in \Cref{eq:general_form_estimator},
  \[ \kappa := \sum_{k=1}^K \KL(Q^k, P^k) + \frac{\|V - V^0\|_F^2}{2\sigma_V^2}, \]
  and
  \[ \hat{L}(Q) := \frac{1}{m} \sum_{(x, y) \in S} \Pr\left\{\argmax \left[\sum_{k=1}^K w^k h^k(x) \right] \ne y \right\} \]
  is the stochastic predictor sample error where the probability is over
  independent draws of \(w^k \sim N(v_k, \sigma_V^2 I), h^k \sim Q^k \) for all
  \(k \in [K]\).
\end{theorem}

\section{Numerical Experiments}\label{section:results}

For numerical evaluation and the tightest possible values of bounds, a few
further ingredients are needed, which are here described. We also give the
specific way these are evaluated in our later experiments.

\paragraph{Bounding the empirical error term.} We note that there is rarely a
closed form expression for \(\hat{L}(Q)\), as there is in the binary SHEL bound. In the multi-class bounds, this term must be estimated and
bounded by making many independent draws of the parameters and using the fact
that the quantity is bounded in \([0, 1]\) to provide a concentration bound
through, for example, Hoeffding's inequality. This adds a penalty to the bound
which reduces with the number of independent draws and thus the amount of
computing time invested in calculating the bound, but this is not a theoretical
drawback of the bound. We give here a form which is useful in the neural network
setting, where it is computationally efficient to re-draw predictors for every
prediction, but we make \(T\) passes through the dataset to ensure a tight
bound. This formulation is considerably more computationally efficient than
drawing a single \(h\) for every pass of the dataset.

\begin{theorem}[Train Set Bound]\label{theorem:train-bnd}
  Let \(Q\) be some distribution over predictors and \(h^{i, t} \sim Q\) be
  i.i.d. draws for \(i \in [m], t \in [T]\). Then with probability at least
  \(1 {-} \delta'\),
  \[ \hat{L}(Q) \le \frac{1}{mT} \sum_{i=1}^m \sum_{t=1}^T \ell(h^{i, t}(x_i), y_i) + \sqrt{\frac{\log\frac{1}{\delta'}}{2mT}}.\]
\end{theorem}

In our results, we will set \(\delta' = 0.01\) (zero in the binary SHEL case), \(T = 20\), and
the generalisation bound \(\delta = 0.025\); combining them our overall results
will hold with probability at least \(\delta + \delta' = 0.035\), as in
\citet{dziugaite2017computing}.

\paragraph{Variance Parameters \(\beta\) and \(\sigma\).} The parameters
\(\beta\), \(\sigma_V\) and \(\sigma_U\) control the variances of the weights in
the stochastic estimator defined by \(Q\), but fulfil different functions. The
\(\beta\) parameter appears in the non-stochastic shallow network \(F_{U, V}\)
and thus affects the final predictions made and the training by SGD, and can be
related to data normalisation as discussed above. We therefore set it to the
fixed value of \(\beta = 5\) in all our experiments.

However the \(\sigma\) parameters appear only on the right hand side of the
bounds for multi-class SHEL and GELU, and can be tuned to provide the tightest bounds---as
they grow the KL term reduces but the performance of \(Q\) will degrade. We
therefore optimise the final bounds over a grid of \(\sigma\) values as follows:
choose a prior grid of \(\sigma_V\) values,
\(\sigma_V \in \{\sigma_V^1, \dots, \sigma_V^r\}\), and combine via a union
bound argument to add a \(\log(r)\) term to \(\kappa\) where \(r\) is the number
of grid elements. The same practice is applied to \(\sigma_U\) in the GELU case.
In practice we use a grid \(\sigma \in \{0.05, 0.06, \dots, 0.2\}\) for both.
Thus the tuning of \(\sigma_U\) and \(\sigma_V\) is not a feature of the bound like \(\beta\), but rather a tool to optimise the tightness of the bounds.

The parameter \(T\) appearing in \Cref{theorem:binary} fufils a similar
function, trading off the performance of \(\hat{L}(Q^{\otimes T})\) versus the
complexity term, but we do not optimise it like the above in our experiments, fixing it to
\(T = 500\) in all our results.

\paragraph{Coupling Procedure.} We adopt a \(60\%\)-prefix coupling procedure for
generating the prior weights \(U^n, V^n\) (rather than \(U^0, V^0\), and similarly in the binary case) as in
\citet{DBLP:journals/corr/abs-2006-10929}. This works by taking the first
\(60\%\) of training examples used in our original SGD run and looping them in
the same order for up to \(4000\) epochs. Note that this also replaces \(m\) by \(m - n\) and \(S\) by \(S^{\operatorname{bnd}}\) in the bounds, so we are making a trade off between optimising the prior and the tightness of the bound (affected by \(m - n\)). These are used to train a prior model
of the same architecture with the same learning rate from the same
initialisation (this is valid because the initialisation is data-independent).
The best bound from the generated prior weights was chosen (with a small penalty
for this choice added to the bound via a union argument).

\paragraph{Numerical Results.} In order to evaluate the quality of the bounds
provided, we made many evaluations of the bound under many different training
scenarios. In particular we show that the bound behaves in similar ways to the
test error on changes of the width, learning rate, training set size and random
relabelling of the data.

The following results follow by training \(\beta\)-normalised SHEL and GELU
networks with stochastic gradient descent on the cross-entropy loss to a
fixed cross entropy value of \(0.3\) for Fashion-MNIST and \(0.1\) for MNIST.
When evaluating the binary SHEL bound (\Cref{theorem:binary}) we use binarised versions
of the datasets
where the two classes consist of the combined classes \(\{0, \dots, 4\}\) and \(\{5, \dots, 9\}\)
respectively (following \citealp{dziugaite2017computing,NIPS2019_8911}), training to cross-entropy values of \(0.2\) for Bin-F (binarised Fashion-MNIST) and \(0.1\) for Bin-M (binarised MNIST) respectively.
We trained using SGD with momentum \(=0.9\) (as suggested by
\citealp{DBLP:journals/corr/HendrycksG16} and following
\citealp{DBLP:journals/corr/abs-2107-03955}) and a batch size of \(200\), or
without momentum and a batch size of \(1000\) (with this larger batch size
stabilising training).
We evaluated for ten different random seeds, a grid search of learning rates
\(\in \{0.1, 0.03, 0.01\}\) without momentum, and additionally
\(\in \{0.003, 0.001\}\) with momentum (where small learning rate convergence
was considerably faster), and widths \(\in \{50, 100, 200, 400, 800, 1600\}\) to
generate the bounds in \Cref{table:best-results}.

\begin{table}[h!]

\begin{center}

  \emph{Best Coupled Bounds with Momentum}
\begin{tabular}{lcccc}
  \toprule
             & Data    & Test Err & Full Bnd  &  Coupled Bnd \\
  \midrule
SHEL        & Bin-M    & 0.038      & 0.837  &   0.286 \\
SHEL        & Bin-F    & 0.085      & 0.426  &   0.297 \\
SHEL        & MNIST    & 0.046      & 0.772  &   0.490 \\
SHEL        & Fashion  & 0.150      & 0.984  &   0.727 \\
GELU        & MNIST    & 0.043      & 0.693  &  0.293  \\
GELU        & Fashion  & 0.153      & 0.976  &  0.568  \\
\bottomrule
\end{tabular}

\vspace{1em}

  \emph{Best Coupled Bounds without Momentum}
\begin{tabular}{lcccc}
  \toprule
             & Data    & Test Err & Full Bnd  &  Coupled Bnd \\
  \midrule
SHEL        & Bin-M    &  0.037  &  0.835  &  0.286 \\
SHEL        & Bin-F    &  0.085  &  0.425  &  0.300 \\
SHEL        & MNIST    &  0.038  &  0.821  &  0.522 \\
SHEL        & Fashion  &  0.136  &  1.109  &  0.844 \\
GELU        & MNIST    &  0.036  &  0.742  &  0.317 \\
GELU        & Fashion  &  0.135  &  1.100  &  0.709 \\
\bottomrule
\end{tabular}

\end{center}

\caption{Results for \(\beta\)-normalised (with \(\beta=5\)) SHEL and GELU networks trained
  with and without momentum SGD on MNIST, Fashion-MNIST and binarised versions of the above, after a grid search
  of learning rates and widths as described above. Results shown are those
  obtaining the tightest coupled bound (calculated using \Cref{theorem:shel-bnd} and \Cref{theorem:gelu-bnd} for the multi-class datasets, and \Cref{theorem:binary} for the binary datasets), with the accompanying full train set
  bound and test error for the same hyper-parameter settings.}
\label{table:best-results}
\end{table}

\begin{figure}[ht]
  \begin{center}
  \centerline{\includegraphics[width=\columnwidth]{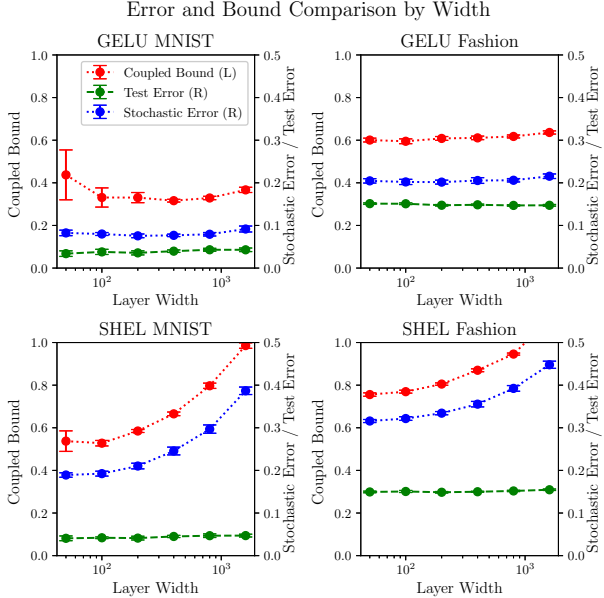}}
  \vskip -0.1in
  \caption{Changes in bound on \textbf{left} (L) hand axis, and test error and
    stochastic bound error \(\hat{L}_{\operatorname{bnd}}(Q)\) on the
    \textbf{right} (R) axis versus width for SHEL and GELU networks trained with
    momentum SGD and learning rate \(0.01\) on Fashion-MNIST and MNIST. Error
    bars show 1 standard deviation from ten different random seeds. The
    different scales are chosen so the trade-off between
    \(\hat{L}_{\operatorname{bnd}}(Q)\) and complexity terms can be seen
    more easily by neglecting the overall factor of \(2\), and the trends can be
    seen more clearly. We include an option in our code to generate these
    figures with a common scaling instead.}
    \label{fig:width-all}
  \end{center}
\vskip -0.4in
\end{figure}

From these results we also show plots in \Cref{fig:width-all} of the test error,
stochastic error \(\hat{L}_{\operatorname{bnd}}(Q)\) and best prior bound
versus width for the different dataset/activation combinations, with more plots
given in the appendix. We also note here that in all except the width \(=50\)
case, our neural networks have more parameters than there are train data points
(\(60000\)).
Using the test set, we also verified that assumption (\(\star\)) holds in all cases in which it is used to provide bounds.

\begin{figure}[ht]
  \begin{center}
  \centerline{\includegraphics[width=\columnwidth]{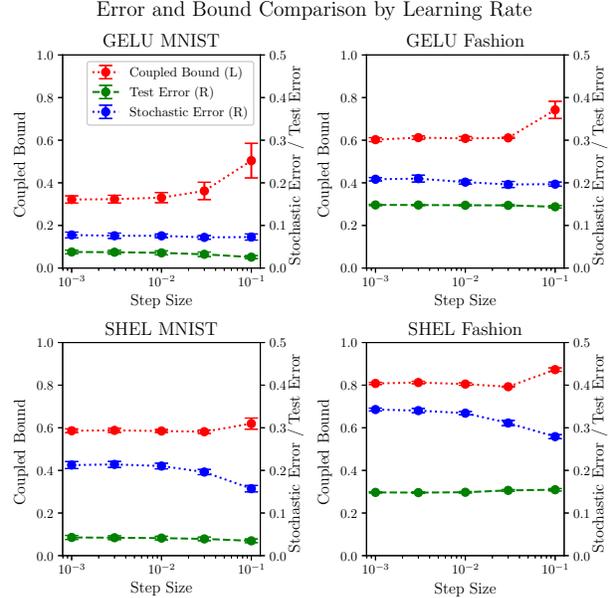}}
  \vskip -0.1in
  \caption{Changes in bound on \textbf{left} (L) hand axis, and test error and stochastic
    bound error \(\hat{L}_{\operatorname{bnd}}(Q)\) on the \textbf{right} (R) axis versus
    learning rate for width 200 SHEL and GELU networks trained with momentum SGD
    on Fashion-MNIST and MNIST. Scales are as in \Cref{fig:width-all}.}
    \label{fig:lrs-all}
  \end{center}
\vskip -0.4in
\end{figure}

\section{Discussion}\label{section:discussion}

In \Cref{table:best-results} we have given the first non-vacuous bounds for two
types of deterministic neural networks trained on MNIST and Fashion-MNIST
through a standard SGD learning algorithm, both with and without momentum. The
coupled bounds are in all cases far from vacuous, with even the full bounds
being non-vacuous in most cases, particularly on the easier MNIST task. Further,
\Cref{fig:width-all,fig:lrs-all} show that the bounds are robustly non-vacuous
across a range of widths and learning rates. Since these are direct bounds on
\(L(F_{U, V})\) rather than the usual PAC-Bayes \(L(Q)\), we emphasise that (for
fixed hyper-parameters) no trade off is made between the tightness of the bound
and the real test set performance, which is usually worse for a higher-variance
(and thus more tightly bounded) \(Q\).

\paragraph{Stability and Robustness Trade-Off.} The two main contributions to the
bound are the empirical error \(\hat{L}(Q)\) and the KL divergence incorporated
in \(\kappa\). \(\hat{L}(Q)\) can be seen roughly as measuring a combination of
the difficulty of the task for our predictor \(F_{U, V}\) combined with some
kind of perturbation resistance of its weights (like the idea of a flat minimum
originated in \citealp{DBLP:conf/colt/HintonC93} and discussed at length by
\citealp{dziugaite2017computing}); while \(\kappa\) is here an empirical measure
of the stability of the training method, scaled by the inverse width of the
perturbation robustness.

When optimising the trade-off between these terms through a choice of
\(\sigma_U, \sigma_V\) values, we find that the complexity contribution to the
bound remains relatively consistent across datasets and architectures, while it
is the stochastic error that varies. This is especially true of multi-class SHEL networks as
seen in \Cref{fig:width-all}, perhaps since there is no easy way to set the
stochastic error small by adjusting the variability of the \(Q\) hidden layer.
This is in direct contrast to many works
\citep{DBLP:conf/iclr/JiangNMKB20,DBLP:conf/nips/DziugaiteDNRCWM20} evaluating
the predictive ability of PAC-Bayesian bounds for generalisation on
hyper-parameter changes, which fix the weight variances as the largest leading
to a bound on \(\hat{L}(Q)\) of a fixed value, say \(0.1\). Our results show
that this approach may be sub-optimal for predicting generalisation, if as in
our results the optimal trade-off tends to fix the \(\kappa\) term and trade off
the size of \(\hat{L}(Q)\) instead of the reverse\footnote{The use of
  bi-criterion plots as suggested by \citet{DBLP:conf/nips/NeyshaburBMS17} may
  therefore offer an better alternative when comparing vacuous bounds.}.

\paragraph{Width Comparison.} For the width comparisons we note that it is
difficult to discern the real trend in the out-of-sample error of our trained
networks. The test sets only have \(10000\) examples and thus any test-set
estimate of \(L(F_{U, V})\) is subject to error; if the differences between test
errors of two networks of different widths is smaller than about \(0.02\)
(obtained through a Hoeffding bound) it is not possible to say if generalisation
is better or worse. It is therefore possible that the pattern of weaker bounds
for wider SHEL networks seen is a strong amplification of an existing trend, but it
seems more likely it is an artefact of the bound shared with that of
\citet{DBLP:journals/corr/abs-2107-03955}.
Assuming the latter conclusion that the trained network true error really is
relatively width-independent, the GELU bound does better matching this
prediction (with this also being true in the momentum-free case, see appendix).
The value of \(\hat{L}_{\operatorname{bnd}}(Q)\) stays roughly constant as
width increases, while we observe that the optimal bound \(\sigma_U\) tends to
decrease with increasing width. We attribute to this the tighter bounds for wide
GELU networks, since the SHEL network has no comparable way to reduce the
randomness of the hidden layer in \(Q\), as we discuss at the end of
\Cref{section:gelunet}.

\paragraph{Lower-Variance Stochastic Predictions.}
Following from the above, we note that in general
\(\hat{L}_{\operatorname{bnd}}(Q)\) is smaller for comparably-trained GELU
networks than the SHEL networks. We speculate that this arises from the
increased randomness of the hidden layer of \(Q\) in \Cref{theorem:shel-bnd}:
the sign activation is only \(\{+1, -1\}\)-valued and the amount of information
coming through this layer is therefore more limited; and a \(\{+1, -1\}\)-valued
random variable has maximum variance among \([+1, -1]\)-bounded variables of
given mean. In future work we will explore whether variance reduction techniques
such as averaging multiple samples for each activation can improve the tightness
of the bounds, but we also emphasise both that the bounds are still non-vacuous
across a range of widths, and that the ability to adjust this variability is a
central advantage of our new GELU formulation.

\paragraph{Learning Rate Comparison and Stability.} In the case of training with
momentum SGD we see that a very large learning rate leads to weaker and
higher-variance bounds, with significantly larger norm contribution in
\(\kappa\). We speculate this arises because of the reduced stability at such high
rates: we found in general that small batch sizes (particularly under vanilla
SGD) and fast learning rates caused the training trajectory of \(U^n, V^n\) to
diverge more greatly from that of \(U, V\).

\paragraph{Improving Prior Coupling.} With the instability of high learning rates
and the empirical observation that in many cases \(\hat{L}(Q)\) was very close
to \(L(Q)\) (as estimated from the test set), we see that there is a degree of
slackness in the bound arising from the \(\kappa\) term. We speculate that it
may be possible to make more efficient use of the sample \(S\) in constructing
\(U^n, V^n\) to reduce this term further. This might be possible through an
improved coupling scheme, or through extra side-channel information from
\(S^{\operatorname{bnd}}\) which can be compressed (as per
\citealp{zhou2018nonvacuous}) or is utilised in a differentially-private manner
(as by \citealp{DBLP:conf/nips/Dziugaite018}).

\paragraph{Majority Votes.} In our results we rely on the novel idea of
randomised single-hidden-layer neural networks as the expectation or majority
vote of randomised predictors for de-randomisation of our PAC-Bayes bound.
For the multi-class bounds we rely on an additional assumption, so a first step in future work could be providing further conditions under which this assumption can be justified without relying on a test set.
Next, we
found empirically (similarly to many PAC-Bayesian works) that
\(L(Q) > L(F_{U, V})\), in other words the derandomised predictor was better
than the stochastic version on the test set. By de-randomising through the first
order bound, we introduce a factor of \(2\) which cannot be tight in such cases.
Removal of this term would lead to considerably tighter bounds and even
non-vacuous bounds for CIFAR-10 \citep{Krizhevsky09learningmultiple}, based on
preliminary experiments, where the training error for one-hidden-layer networks
on CIFAR-10 was greater than \(0.5\) so such bounds could not be non-vacuous,
but the final bounds were only around \(1.1{-}1.2\).
Improved bounds for the majority
vote have been the focus of a wide variety of PAC-Bayesian works
\citep{DBLP:conf/nips/LacasseLMGU06,DBLP:conf/nips/MasegosaLIS20}, and
can theoretically give tighter results for \(L(\MV_Q)\) than \(L(Q)\), but
these are not yet competitive. They universally led to inferior or vacuous
results in preliminary experiments. However, there is still much scope for
exploration here: alternative formulations of the oracle C-bound lead to
different empirical bounds, and improvement of the KL term (which appears more
times in an empirical C-bound than \Cref{theorem:seeger}) may improve these
bounds more than the first order one. We also hope that offering this new
perspective on one-hidden-layer networks as majority votes can lead to better
understanding of their properties, and perhaps even of closely-related Gaussian
processes \citep{Neal1996}.

\paragraph{Deeper networks and convolutions.}
An extremely interesting question whether this approach will generalise to
convolutions or deeper networks. For convolutions, the parameter sharing is not
a problem as separate samples can be taken for each convolution kernel position
(although potentially at a large KL divergence cost that might be mitigated
through the use of symmetry). For deeper networks the answer is less clear, but
the empirically-observed stability of most trained networks to weight
perturbation would suggest that the mode of a Bayesian neural network may at
least be a close approximation to its majority vote, a connection that could
lead to further results.

\paragraph{Summary.} We have provided non-vacuous generalisation bounds for
shallow neural networks through novel methods that make a promising new link
to majority votes. Although some aspects of our approach have recently appeared in the
PAC-Bayesian literature on neural networks, we note that all previous results
obtaining non-vacuous generalisation bounds only apply to randomised
versions of neural networks. This often leads to degraded test set performance
versus a deterministic predictor. By providing bounds directly on the
deterministic networks we provide a setting through which the impact of
robustness, flat-minima and stability on generalisation can be explored
directly, without making potentially sub-optimal trade-offs or invoking stringent
assumptions.

In future work we intend to address two main potential sources of improvement:
through progress in majority votes to tighten the step from stochastic to deterministic predictor; and
through development of the prior (perhaps thorough improved utilisation of data), a
strand running parallel to much PAC-Bayesian research on neural networks.

\subsection*{Acknowledgements}

F.B. acknowledges the support of the EPSRC grant EP/S021566/1.
B.G. acknowledges partial support by the U.S. Army Research Laboratory, U.S. Army Research Office, U.K. Ministry of Defence and the U.K. Engineering and Physical Sciences Research Council (EPSRC) under grant number EP/R013616/1; B.G. also acknowledges partial support from the French National Agency for Research, grants ANR-18-CE40-0016-01 and ANR-18-CE23-0015-02.


\bibliography{bibliography}
\bibliographystyle{icml2022}

\newpage
\appendix
\onecolumn


\clearpage

\section{Proofs}

\begin{proof}[Proof of \Cref{theorem:binary}]
  We consider randomised functions
  \(f(x) = \frac{1}{T}\sum_{t=1}^T \sign(w_t \cdot x)\) with
  \(w_1, \dots, w_T \sim Q^{\otimes T}\) identically and independently
  distributed. Here \(Q\) is a mixture of Gaussians distribution with \(2K\)
  components; we denote by \(Q_k = \operatorname{Categ}(q)\) the distribution
  over the choice of component, and by \(Q^k\) the corresponding component. We
  choose the mixture component weights
  \[q = \frac{1}{\|v\|_1}[\max(0, v_1), \dots, \max(0, v_K), \max(0, -v_1), \dots, \max(0, -v_K)],\]
  and component distributions \(Q^k = N(u_k, \frac12 \beta^{-2}I)\) for \(k \in {1, \dots, K}\), and
  \(Q^k = N(-u_k, \frac12 \beta^{-2}I)\) for \(k \in {K+1, \dots, 2K}\). Here \(u_k\) are the rows of \(U\). This
  dimension-doubling trick accommodates the use of negative final-layer weights.

  A PAC-Bayes bound on the above relates to the SHEL network through the
  following. Firstly, it is easy to show that
  \(\E_{f \sim Q^{\otimes T}}f(x) = \frac{1}{\|v\|_1} F(x)\), where \(F\) is the
  SHEL network with parameters \(U, v\) as given above. This follows using the expectation of a mixture followed by using the aggregation of a sign function under a Gaussian weight given in \Cref{eq:erfaggregation}, which gives
  \[ \E_{f \sim Q^{\otimes T}}f(x) = \sum_{k=1}^{K} q_k \erf{\beta \frac{u_k \cdot x}{\|x\|_2}} + \sum_{k=K+1}^{2K} q_k \erf{\beta \frac{-u_k \cdot x}{\|x\|_2}} = \frac{F(x)}{\|v\|_1}\]

  The predictions of this
  SHEL network, \(\sign F(x)\), are equivalent to a majority vote of \(f(x)\),
  since \(\MV(x) = \sign(\E\sign(f(x)))\) is \(1\) if
  \(\E f(x) \propto F(x) \ge 0\) and vice-versa for \(-1\). Therefore the first
  order bound can be used to see that
  \(\ell(F(x), y) \le 2 \E_{Q^{\otimes T}} \ell(f(x), y)\).

  To obtain a PAC-Bayes bound in full, we choose a set of prior weights
  \(U^0, v^0\) to define a prior \(P\) that takes the same structure as \(Q\).
  The index distribution \(P^k = \operatorname{Categ}(p)\) with
  \[ p = \frac{1}{2\|v^0\|} [|v^0_1|, \dots, |v^0_K|, |v^0_1|, \dots, |v^0_K|],\]
    and component distributions defined as per \(Q^k\) but with weights \(u_k^0\)
  instead.

  Then, using the chain rule for KL divergence \citep{Cover2006} twice,
  \begin{equation}\label{eq:kl-mixture}
    \KL(Q, P) \le \KL(Q_{w,k}, P_{w,k}) \le \KL(Q_{w|k}, P_{w|k}) + \KL(Q_k, P_k)
  \end{equation}
  where \(Q_{w,k}\) and \(Q_{w|k}\) are the joint and conditional distributions
  on \(w\) and mixture index \(k\) (and analogously for \(P\)), as opposed to
  \(Q\), which is a marginal on \(w\).

  Using the definitions of the KL divergence for categorical and Gaussian
  distributions in the above, \(\KL(Q, P)\) is bounded by
  \[ \sum_{k=1}^K q_k \beta \|u_k - u_k^0\|_2^2 + \sum_{k=1}^K q_k \log \frac{q_k}{p_k}  =  \kappa. \]

  Combining \Cref{theorem:seeger} with the fact that
  \(\KL(Q^{\otimes T}, P^{\otimes T}) = T \KL(Q, P)\) since the \(T\) copies are
  i.i.d., the following holds with probability \(\ge 1 - \delta\)
  \[
    L(F_{U, v}) \le 2 \, \smallkl^{-1}\left(\hat{L}(Q^{\otimes T}),\; \frac{T\kappa + \log\frac{2\sqrt{m}}{\delta}}{m}  \right).
  \]

  To complete the result we also note the closed form for \(\hat{L}(Q^{\otimes T})\) given
  through the following. The average misclassification loss
  \begin{align*}
    \E_{Q^{\otimes T}}\ell(f(x), y) &= \Pr_{Q^{\otimes T}}\left(y f(x) \le 0 \right) \\
    &= \Pr_{Q^{\otimes T}}\left(\sum_{t=1}^T y \sign(w^t \cdot x) \le 0 \right) \\
    &= \Pr_{Q^{\otimes T}}\left(\sum_{t=1}^T \frac12 (y \sign(w^t \cdot x) + 1) \le \frac12 T \right) \\
    &= \Pr_{Q^{\otimes T}}\left(\sum_{t=1}^T \I_{y = \sign(w^t \cdot x)} \le \frac12 T \right) \\
    &= \operatorname{Bin}\left(\frac{T}{2}; \, T, \Pr_Q(y = \sign(w^t \cdot x))\right) \\
    &= \operatorname{Bin}\left(\frac{T}{2}; \, T, \frac12 \left( 1 + \frac{yF(x)}{\|v\|_1}\right)\right)
  \end{align*}
  where we have interchanged \(\I_{y = \sign(w \cdot x)}  = \frac12 (y \sign(w \cdot x) + 1)\).

  All of the above can be readily extended to the data-dependent prior case,
  replacing \(U^0 \to U^n\), \(v^0 \to v^n\), \(m \to m - n\), and
  \(\hat{L} \to \hat{L}_{\operatorname{bnd}}\).
\end{proof}

\begin{proof}[Proof of \Cref{theorem:main-general}]
  We are considering a distribution on functions of the form
  \(\sum_k w^k h^k(x)\) where for each index \(k \in [K]\) we have
  \(w_k \sim N(\frac{1}{\sigma_V} v_k, I)\) and \(h_k \sim Q_k\). This slightly
  different formulation can take advantage of the scaling-invariance of the
  final layer to the misclassification loss when \(V^0 = 0\), so we can then
  choose \(\sigma_V > 0\) arbitrarily. The expectation of this takes the form
  given in \Cref{eq:general_form_estimator} scaled by \(1/\sigma_V\) and
  leads to the empirical loss above.

  Given another distribution \(P\) taking a similar form with
  \(w_k \sim N(\frac{1}{\sigma_V} v_k^0, I)\) and components \(P_k\), the KL divergence
  can be expressed (using the chain rule for KL divergence) as
  \[ \KL(Q, P) \le \sum_{k=1}^K \KL(Q^k, P^k) + \frac{\|V - V^0\|^2_F}{2\sigma_V^2} .\]

  We prove the overall bound by combining \Cref{theorem:seeger} with the assumption (\(\star\)).
\end{proof}

\begin{proof}[Proof of \Cref{theorem:shel-bnd}]
  Apply the bound from \Cref{theorem:main-general} with the individual units as \(h_k(x) = \sign(w_k \cdot x)\) and
  \(w_k \sim N(u_k, \frac12 \beta^{-2} I)\) alongside \Cref{theorem:main-general}.
  The aggregated form of the sign activation function is given
  in \eqref{eq:erfaggregation}. The prior takes the same form as the posterior
  with weight means \(U^0, V^0\) and the same variances, leading to the form of
  KL divergence for Gaussian weights given in \(\kappa\).
\end{proof}

\begin{proof}[Proof of \Cref{theorem:gelu-bnd}]
  The proof takes the same form as that of \Cref{theorem:shel-bnd}. We
  note that the expectation under the given probability distributions of
  \(\E [W_2 (\I_{W_1 x} \otimes (W_1' x))] = \|x\|_2 F^{\operatorname{GELU}}_{U, V}(x)\), but since the
  misclassification loss is scaling-invariant this gives equivalent results.
  Choosing appropriate prior forms as in \Cref{theorem:shel-bnd} gives the KL
  divergence which we substitute into \Cref{theorem:main-general}.
\end{proof}

\begin{proof}[Proof of \Cref{theorem:train-bnd}]
  Define
  \(\xi = \sum_{i=1}^m \sum_{t=1}^T \frac{1}{mT} \ell(h^{i, t}(x_i), y_i)\)
  which has expectation \(\E_Q \xi = \hat{L}(Q)\). Since this quantity is a
  sum of \(m T\) independent random variables in \(\{0, 1/mT \}\), application
  of Hoeffding's inequality gives the result.
\end{proof}

\section{Additional Results and Code}

We provide all of our results and code to reproduce them along with the figures
(including with the option of using the same scaling for the bound and errors,
as described in \Cref{fig:width-all}) in the supplementary material. We also note here that the ``erf'' function is included in a wide variety of common deep learning libraries.

\begin{figure}[ht]
  \begin{center}
  \centerline{\includegraphics[width=0.45\columnwidth]{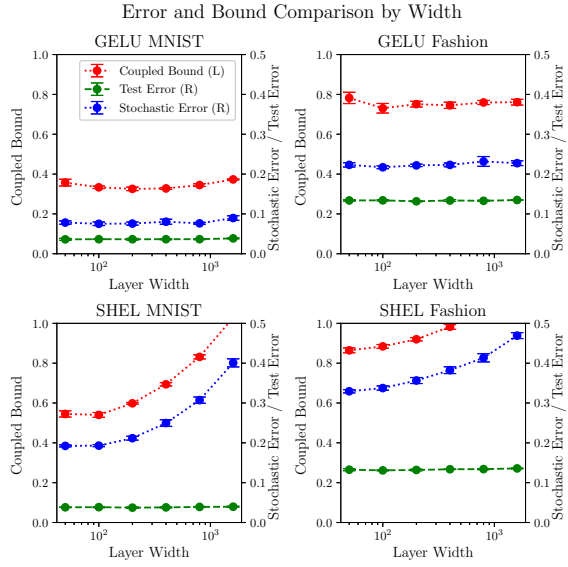}}
  \vskip -0.1in
  \caption{Changes in bound on \textbf{left} (L) hand axis, and test error and stochastic
    bound error \(\hat{L}_{S^{\operatorname{bnd}}}(Q)\) on the \textbf{right} (R) axis versus
    width for SHEL and GELU networks trained with vanilla SGD and learning rate
    \(0.01\) on Fashion-MNIST and MNIST. Scales are as in \Cref{fig:width-all}.}
    \label{fig:width-all-vanilla}
  \end{center}
\vskip -0.4in
\end{figure}

\begin{figure}[ht]
  \begin{center}
  \centerline{\includegraphics[width=0.45\columnwidth]{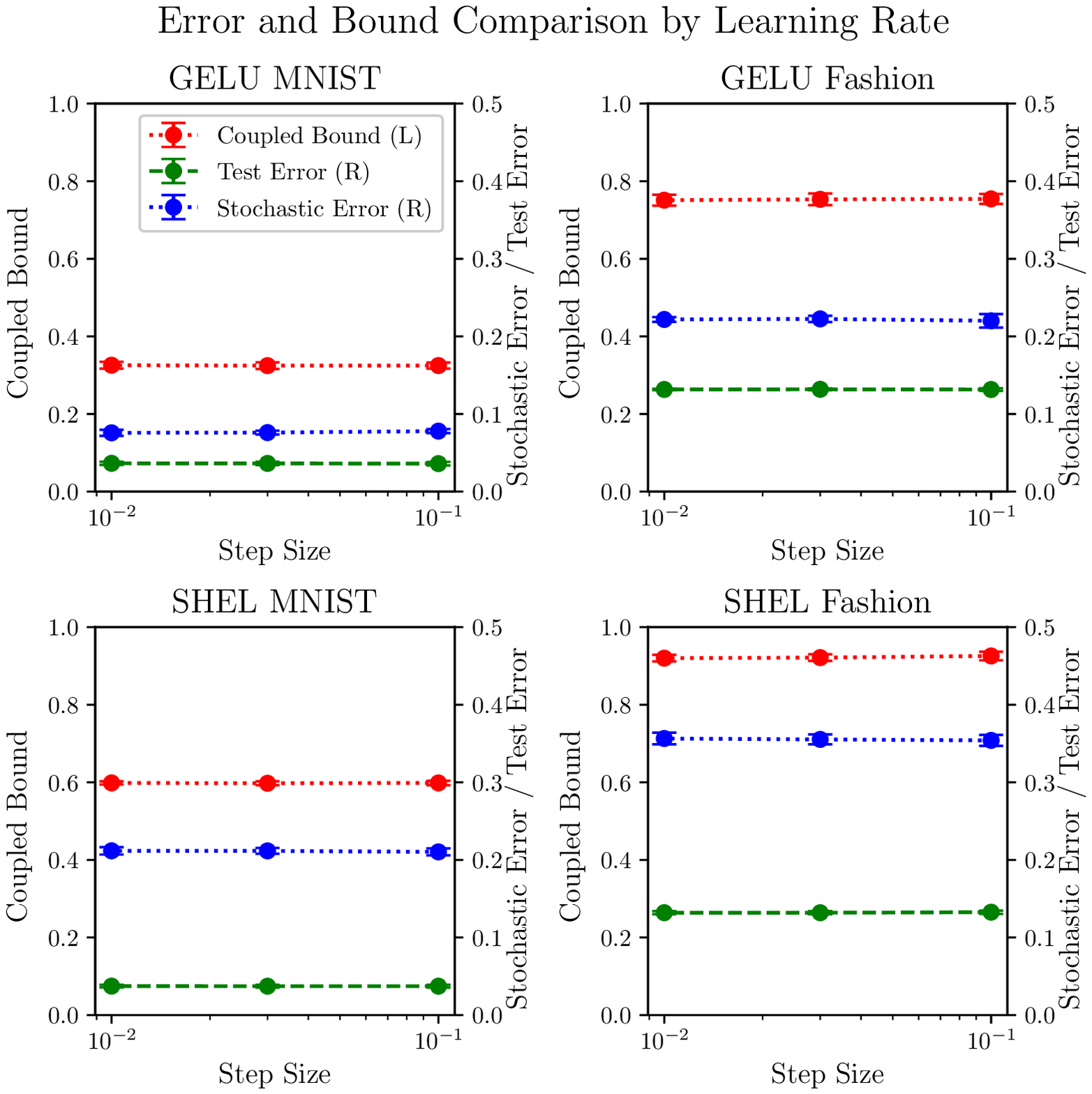}}
  \vskip -0.1in
  \caption{Changes in bound on \textbf{left} (L) hand axis, and test error and stochastic
    bound error \(\hat{L}_{S^{\operatorname{bnd}}}(Q)\) on the \textbf{right} (R) axis versus
    learning rate for width 200 SHEL and GELU networks trained with vanilla SGD
    on Fashion-MNIST and MNIST. Scales are as in \Cref{fig:width-all}.}
    \label{fig:lrs-all-vanilla}
  \end{center}
\vskip -0.4in
\end{figure}

Here we also provide \Cref{fig:width-all-vanilla,fig:lrs-all-vanilla} similar to
\Cref{fig:width-all,fig:lrs-all} for GELU and SHEL networks trained without
momentum and with a batch size of \(1000\), as described in
\Cref{section:results}. We then also provide further similar plots for networks
trained with momentum and a batch size of \(200\) as in \Cref{section:results}
with different learning rates and widths, to show the similar behaviour across a
variety of regimes.

\begin{figure}[h!]
  \centering
  \includegraphics[width=0.8\textwidth]{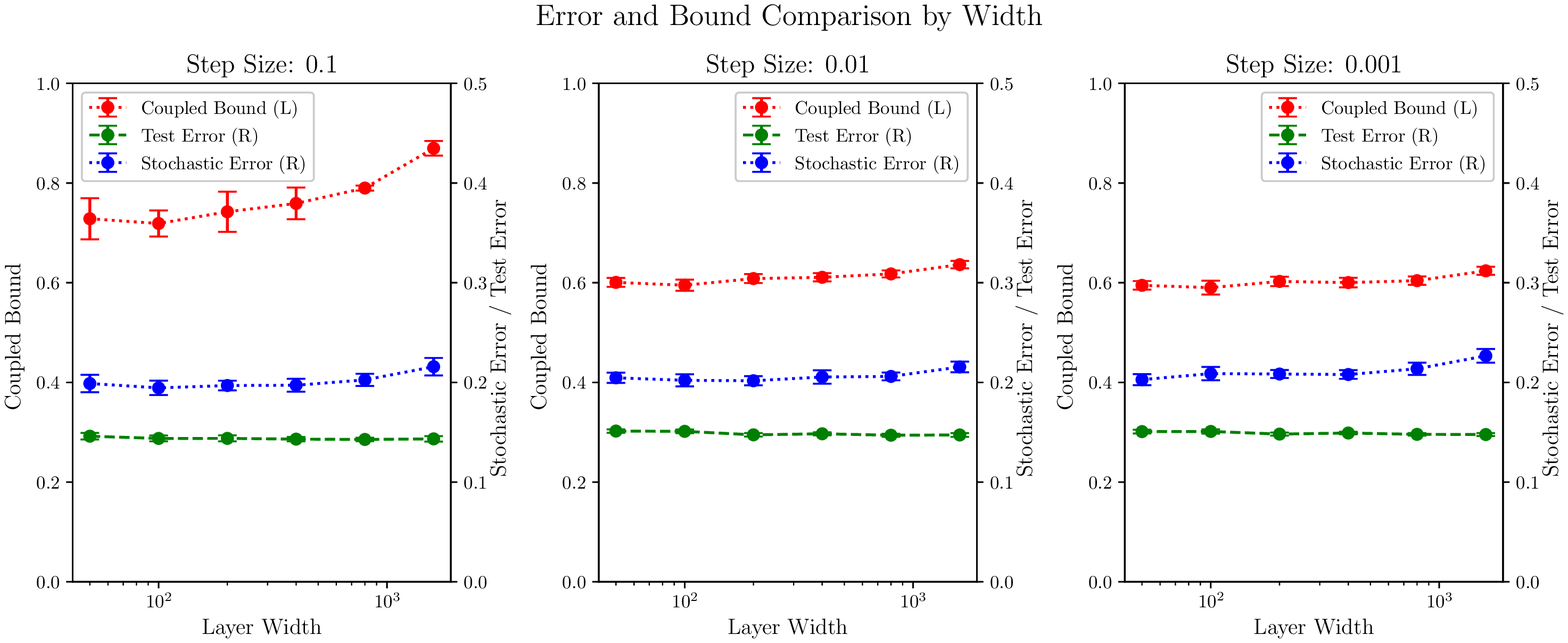}
  \caption{Changes in bound on \textbf{left} (L) hand axis, and test error and
    stochastic bound error \(\hat{L}_{S^{\operatorname{bnd}}}(Q)\) on the
    \textbf{right} (R) axis versus width under fixed other hyperparameters, for
    a GELU network trained with momentum on Fashion-MNIST.}
\end{figure}

\begin{figure}[h!]
  \centering
  \includegraphics[width=0.8\textwidth]{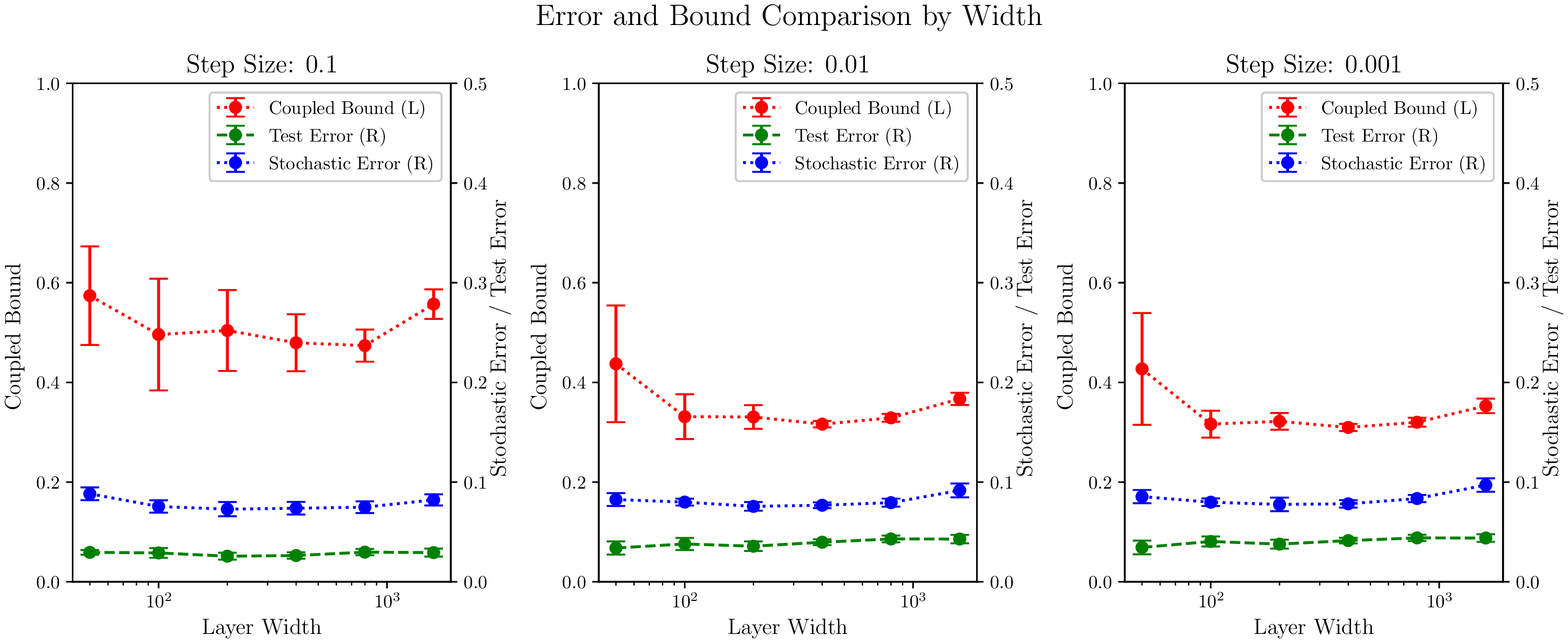}
  \caption{Changes in bound on \textbf{left} (L) hand axis, and test error and
    stochastic bound error \(\hat{L}_{S^{\operatorname{bnd}}}(Q)\) on the
    \textbf{right} (R) axis versus width under fixed other hyperparameters, for
    a GELU network trained with momentum on MNIST.}
\end{figure}

\begin{figure}[h!]
  \centering
  \includegraphics[width=0.8\textwidth]{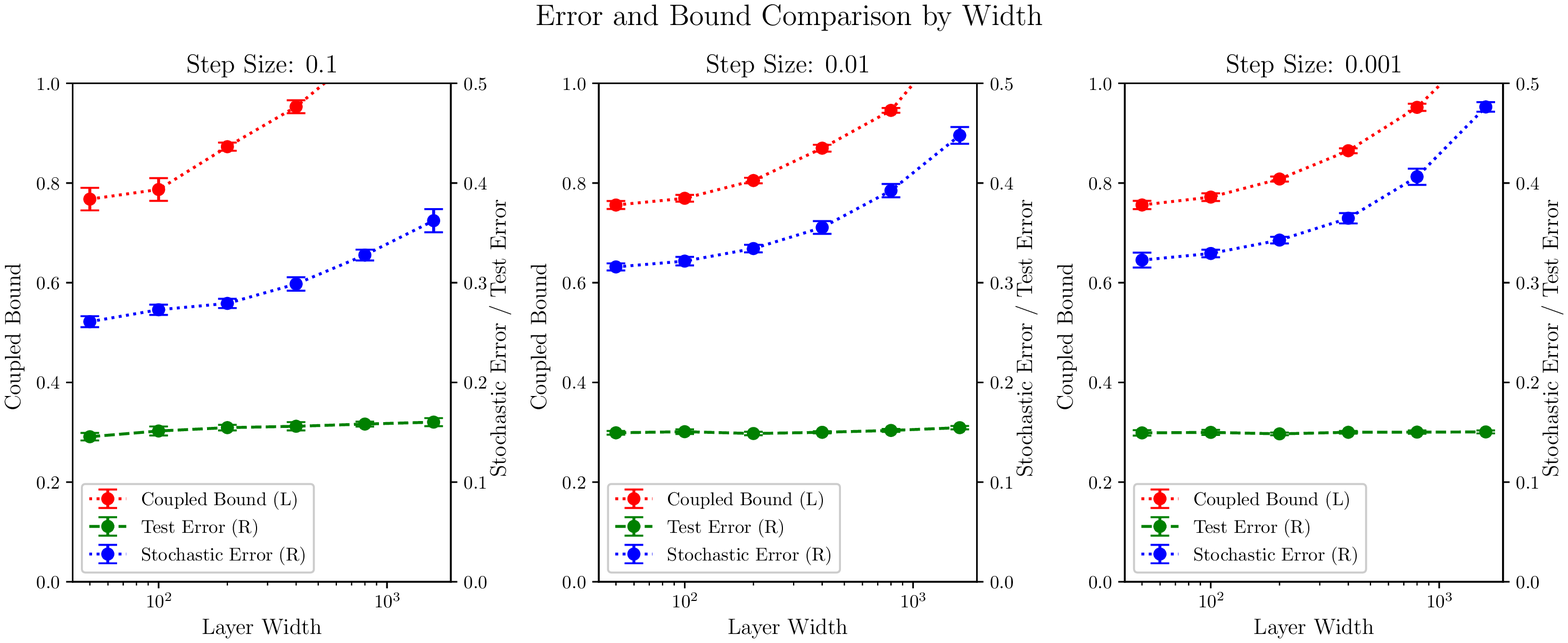}
  \caption{Changes in bound on \textbf{left} (L) hand axis, and test error and
    stochastic bound error \(\hat{L}_{S^{\operatorname{bnd}}}(Q)\) on the
    \textbf{right} (R) axis versus width under fixed other hyperparameters, for
    a SHEL network trained with momentum on Fashion-MNIST.}
\end{figure}

\begin{figure}[h!]
  \centering
  \includegraphics[width=0.8\textwidth]{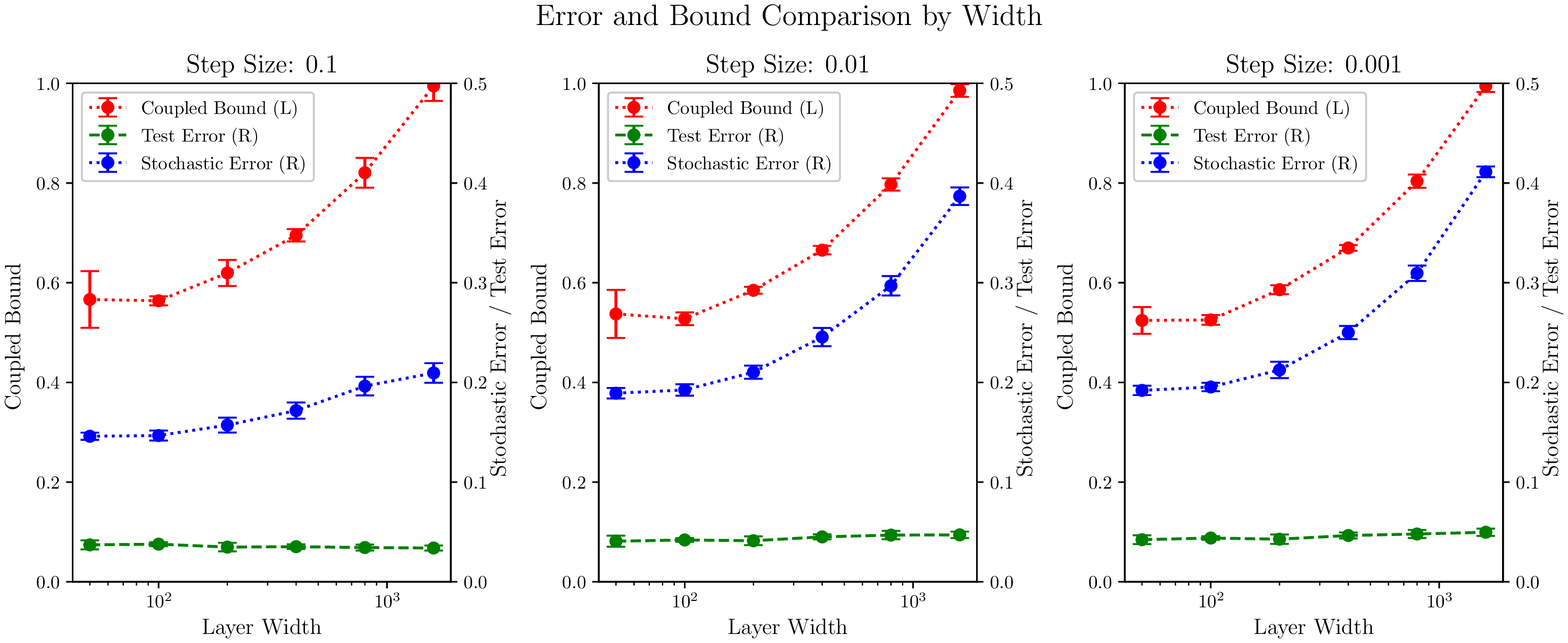}
  \caption{Changes in bound on \textbf{left} (L) hand axis, and test error and
    stochastic bound error \(\hat{L}_{S^{\operatorname{bnd}}}(Q)\) on the
    \textbf{right} (R) axis versus width under fixed other hyperparameters, for
    a SHEL network trained with momentum on MNIST.}
\end{figure}

\begin{figure}[h!]
  \centering
  \includegraphics[width=0.8\textwidth]{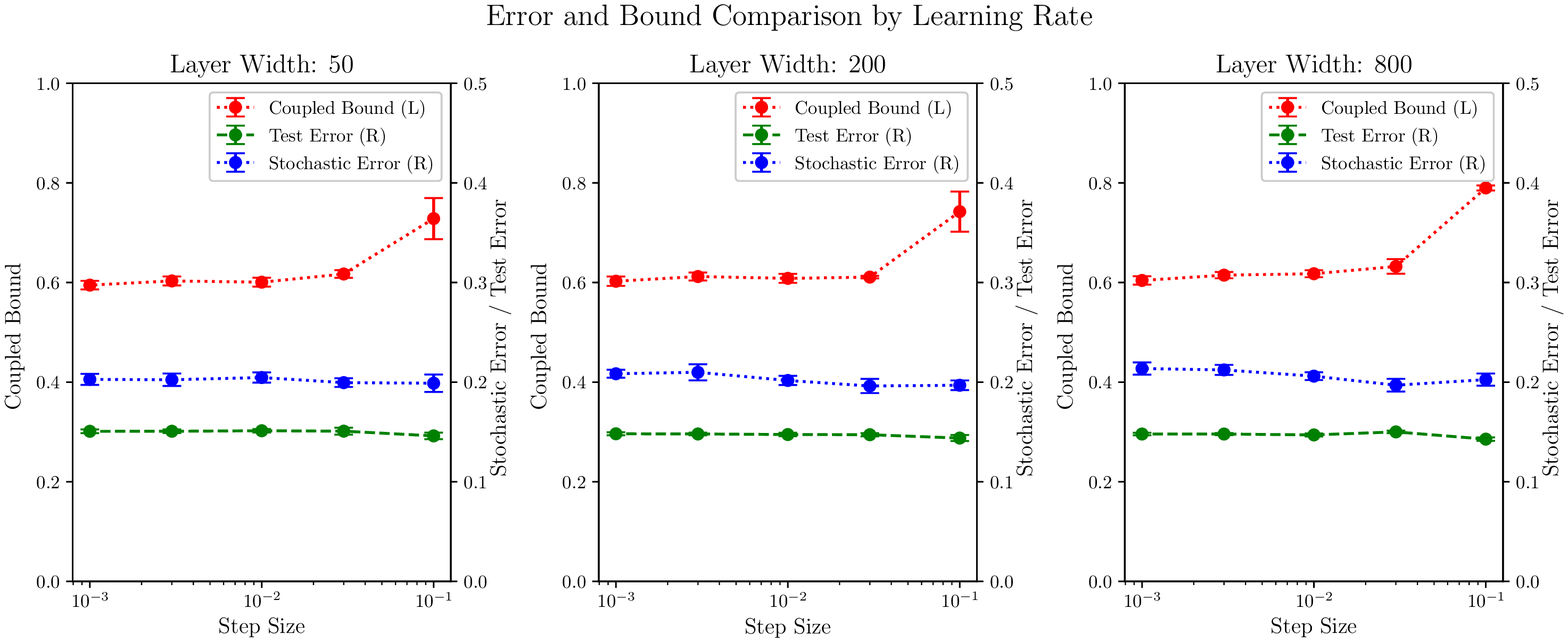}
  \caption{Changes in bound on \textbf{left} (L) hand axis, and test error and
    stochastic bound error \(\hat{L}_{S^{\operatorname{bnd}}}(Q)\) on the
    \textbf{right} (R) axis versus learning rate under fixed other
    hyperparameters, for a GELU network trained with momentum on Fashion-MNIST.}
\end{figure}

\begin{figure}[h!]
  \centering
  \includegraphics[width=0.8\textwidth]{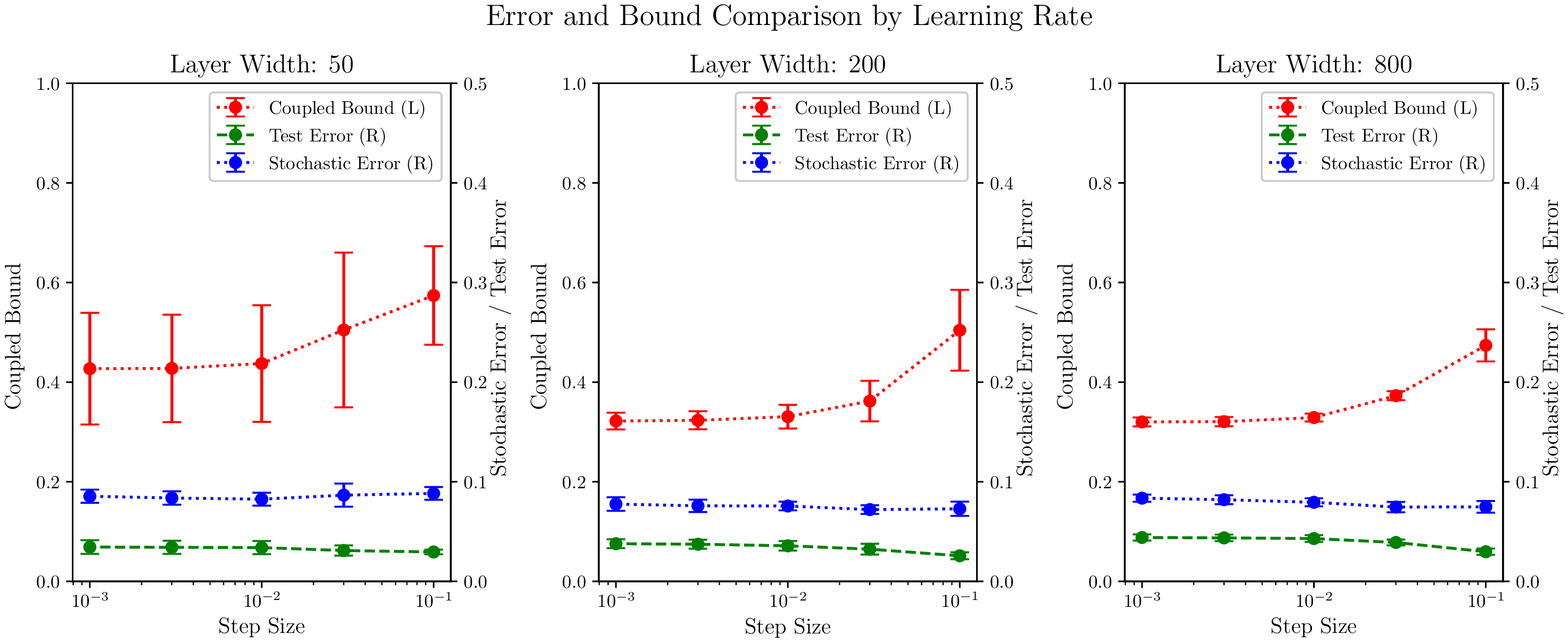}
  \caption{Changes in bound on \textbf{left} (L) hand axis, and test error and
    stochastic bound error \(\hat{L}_{S^{\operatorname{bnd}}}(Q)\) on the
    \textbf{right} (R) axis versus learning rate under fixed other
    hyperparameters, for a GELU network trained with momentum on MNIST.}
\end{figure}

\begin{figure}[h!]
  \centering
  \includegraphics[width=0.8\textwidth]{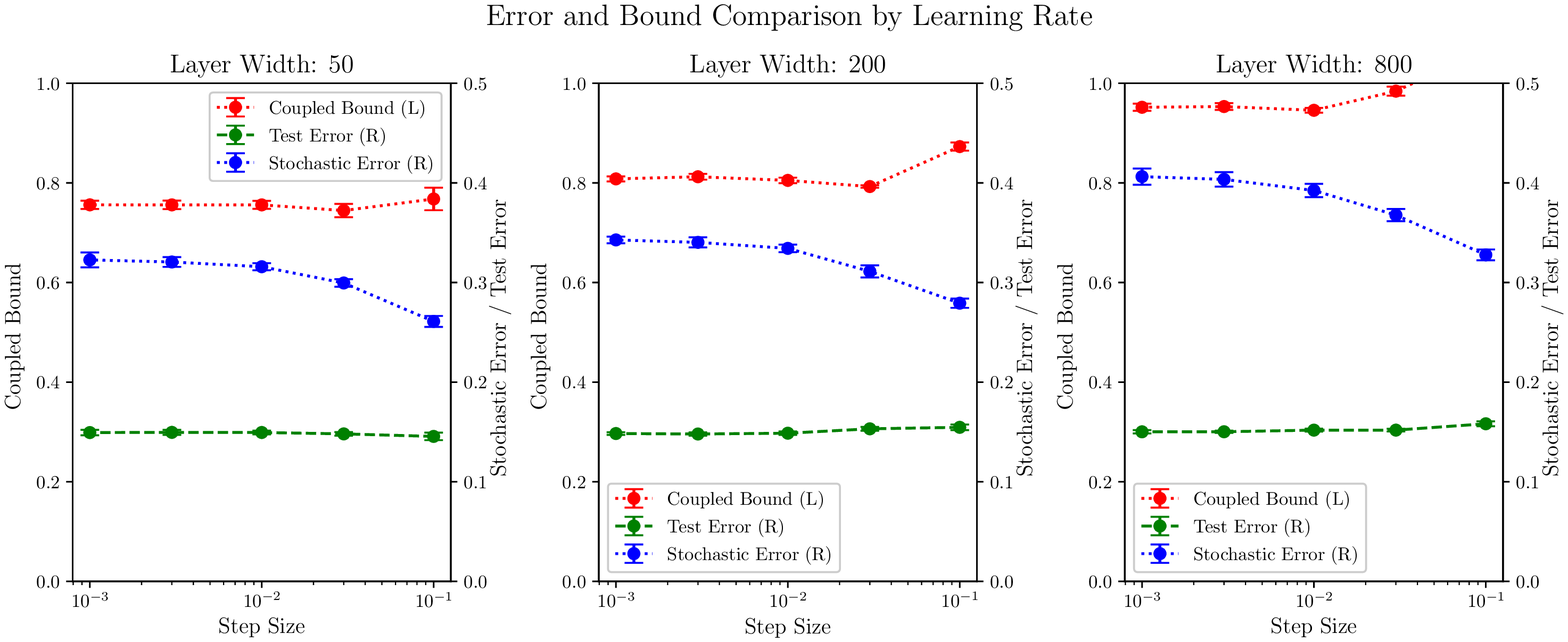}
  \caption{Changes in bound on \textbf{left} (L) hand axis, and test error and
    stochastic bound error \(\hat{L}_{S^{\operatorname{bnd}}}(Q)\) on the
    \textbf{right} (R) axis versus learning rate under fixed other
    hyperparameters, for a SHEL network trained with momentum on Fashion-MNIST.}
\end{figure}

\begin{figure}[h!]
  \centering
  \includegraphics[width=0.8\textwidth]{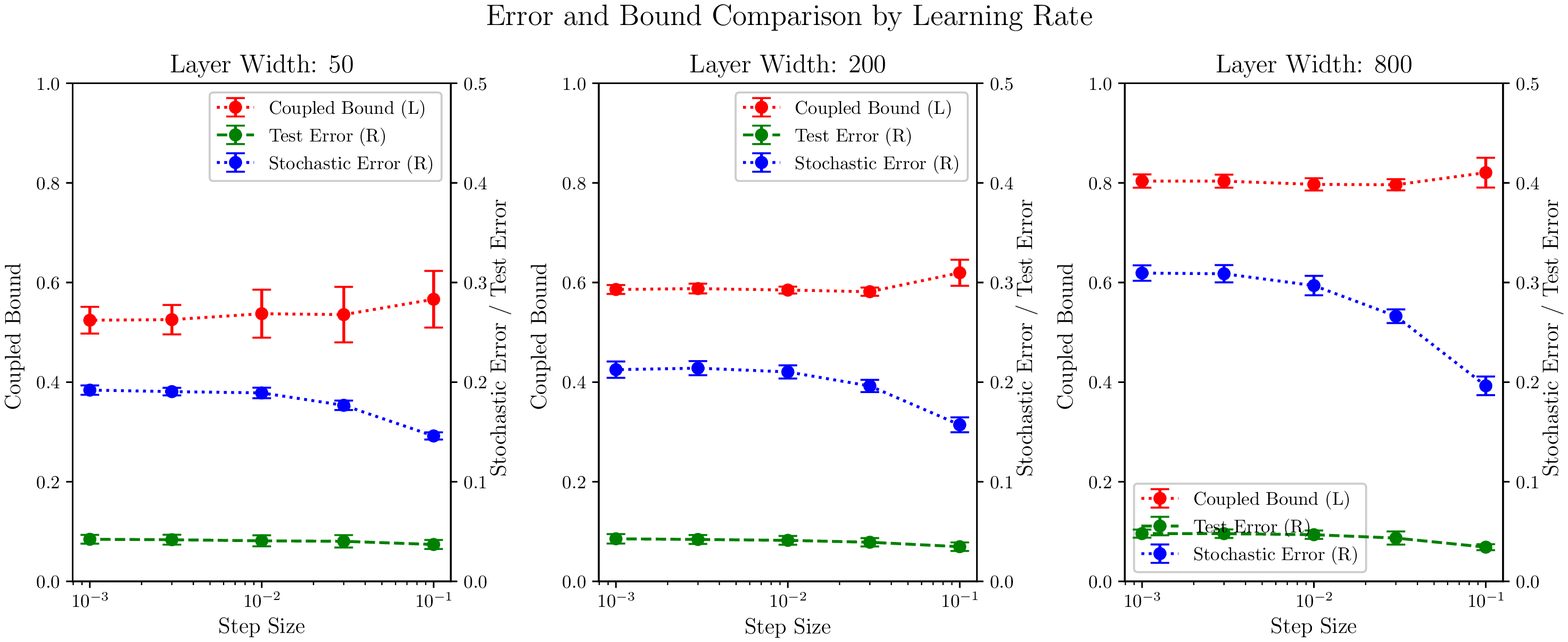}
  \caption{Changes in bound on \textbf{left} (L) hand axis, and test error and
    stochastic bound error \(\hat{L}_{S^{\operatorname{bnd}}}(Q)\) on the
    \textbf{right} (R) axis versus learning rate under fixed other
    hyperparameters, for a SHEL network trained with momentum on MNIST.}
\end{figure}


\end{document}